\documentclass[journal]{IEEEtran}
\usepackage{bm}
\usepackage{comment}
\usepackage{amsmath}
\usepackage{enumerate}
\usepackage{mathrsfs}
\usepackage{makeidx}         
\usepackage{graphicx}        
\usepackage{multicol}        
\usepackage{subfigure}
\usepackage{epsfig,amssymb,latexsym}
\usepackage{psfrag}
\usepackage{algorithmic}
\usepackage{algorithm}

\usepackage{fancyhdr}
\usepackage{multirow}

\usepackage{color}

\renewcommand{\algorithmicrequire}{\textbf{Input:}}   
\renewcommand{\algorithmicensure}{\textbf{Output:}}  

\newcommand{\mb}{\mathbb}
\newcommand{\mr}{\mathrm}
\newcommand{\mc}{\mathcal}

\newcounter{thm_counter}

\newtheorem{theorem}[thm_counter]{Theorem}

\begin{document}
%
\title{Lifelong Metric Learning}
%

\author{Gan~Sun, Yang~Cong,~\IEEEmembership{Senior Member,~IEEE}, Ji Liu, Xiaowei~Xu,~\IEEEmembership{Member,~IEEE}

\thanks{G. Sun is with the State Key Laboratory of Robotics, Shenyang Institute
of Automation, Chinese Academy of Sciences, University of Chinese Academy of Sciences, China, 110016 e-mail: sungan@sia.cn}
\thanks{Y. Cong is with the State Key Laboratory of Robotics, Shenyang Institute
of Automation, Chinese Academy of Sciences, China, 110016 e-mail: congyang81@gmail.com}
\thanks{J. Liu is with the Department of Computer Science, University of Rochester, USA, e-mail: jliu@cs.rochester.edu}
\thanks{X. Xu is with Department of Information Science, University of Arkansas at Little Rock, USA
e-mail: xwxu@ualr.edu}}



\maketitle

\begin{abstract}
The state-of-the-art online learning approaches are only capable of learning the metric for predefined tasks. In this paper, we consider lifelong learning problem to mimic ``human learning'', i.e., endowing a new capability to the learned metric for a new task from new online samples and incorporating previous experiences and knowledge.  Therefore, we propose a new metric learning framework: lifelong metric learning (LML), which only utilizes the data of the new task to train the metric model while preserving the original capabilities. More specifically, the proposed LML maintains a common subspace for all learned metrics, named lifelong dictionary, transfers knowledge from the common subspace to each new metric task with task-specific idiosyncrasy, and redefines the common subspace over time to maximize performance across all metric tasks. For model optimization, we apply online passive aggressive optimization algorithm to solve the proposed LML framework, where the lifelong dictionary and task-specific partition are optimized alternatively and consecutively. Finally, we evaluate our approach by analyzing several multi-task metric learning datasets. Extensive experimental results demonstrate effectiveness and efficiency of the proposed framework.
\end{abstract}




%
\IEEEpeerreviewmaketitle

\section{Introduction}


 \IEEEPARstart{O}{nline} metric / similarity learning has received remarkable success in a variety of applications \cite{weinberger2009distance,zha2009robust,baghshah2009semi}, such as data mining \cite{wang2015survey}, information retrieval \cite{davis2007information} and computer vision \cite{quattoni2009recognizing,luo2014decomposition}, mainly due to its high efficiency and scalability to large-scale dataset. Different from conventional batch learning methods that learn metric model offline with all training samples, online learning aims to exploit one or a group of samples each time to update the metric model iteratively, and is ideally appropriate for tasks in which data arrives sequentially.  

 However, most state-of-the art online metric learning models \cite{jain2009online,chechik2009online,cong2013self} can only achieve online learning from fixed predefined $t$ ($t>0$) metric tasks and cannot add the new task. In this paper, we consider the lifelong learning problem to mimic the ``human learning'', i.e., how to extend the current metric to new tasks while the current functionality of the metric remains. For example, in speech recognition, different people pronouncing the same word differs greatly based on their gender, accent, nationality, or other individual characteristics, and it is highly beneficial to leverage the similarities of datasets from different types of speakers while adapting to the specifics of each particular users. Therefore, the speech recognition library should be delivered to coming speaker's recognition with a set of default speech recognition capabilities, and new speaker-specific metric models need to be added. Another motivating example is in image classification system: a metric learning system can identify whether an image contains an apple or banana, however the user wishes to expand this ability to a new task, e.g., detecting an orange. To achieve this goal, most state-of-the-arts \cite{parameswaran2010large,yang2011multi} should storage training data of all tasks and retrain their models in a time consuming way. Therefore, the key challenge lies on how to learn and accumulate knowledge continuously where early samples are not accessible in the online scenario.

\begin{figure*}[t]
   \subfigure{
       \label{fig:lifelong} 
       \centering
        \begin{minipage}{0.48\textwidth}
         \centering
    \includegraphics[width=250pt, height=130pt]{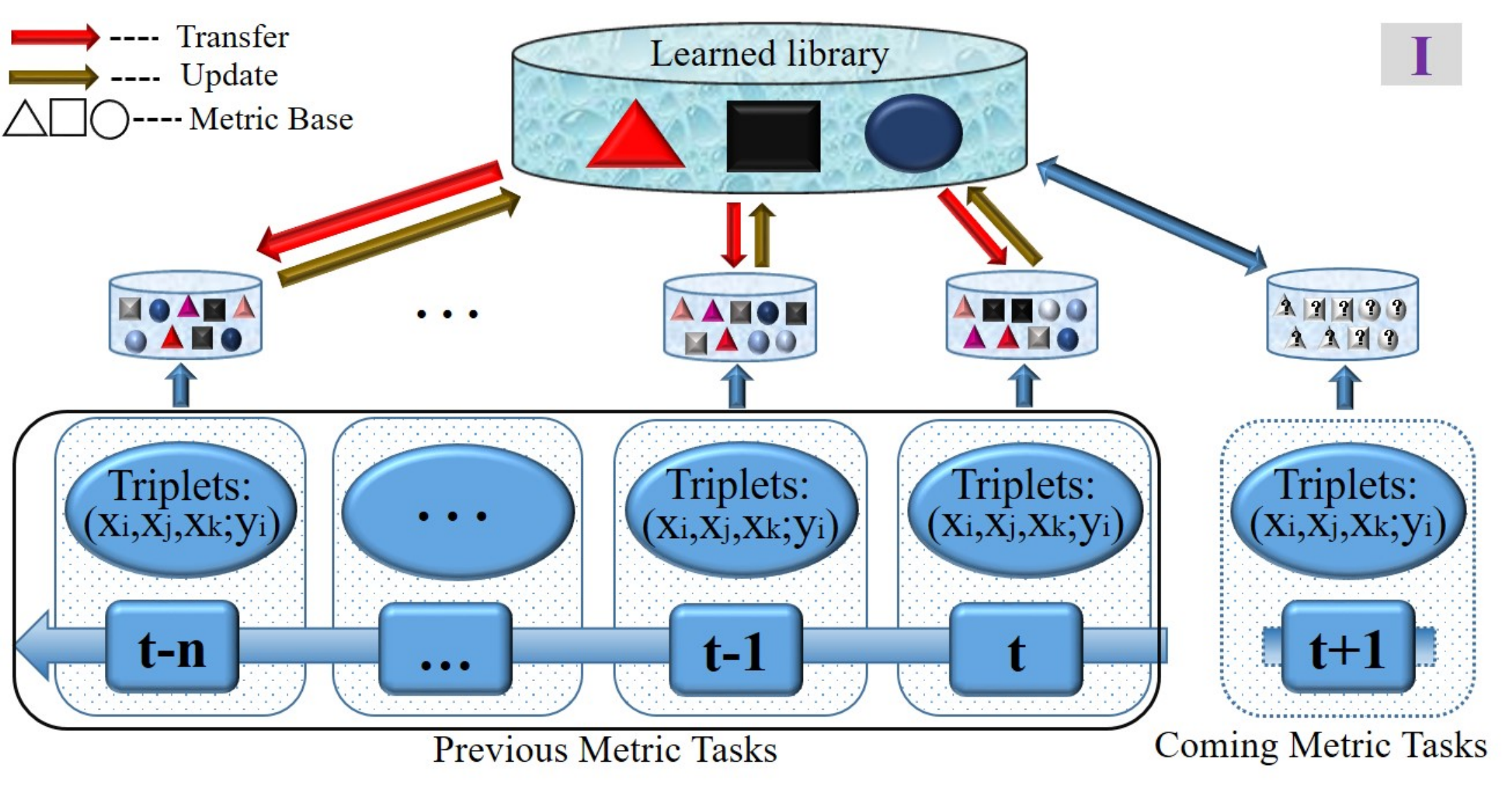}
         \end{minipage}}
         \hspace{0.1mm}
   \subfigure{
       \label{fig:offline} 
        \begin{minipage}[a]{0.48\textwidth}
          \centering
    \includegraphics[width=250pt, height=130pt]{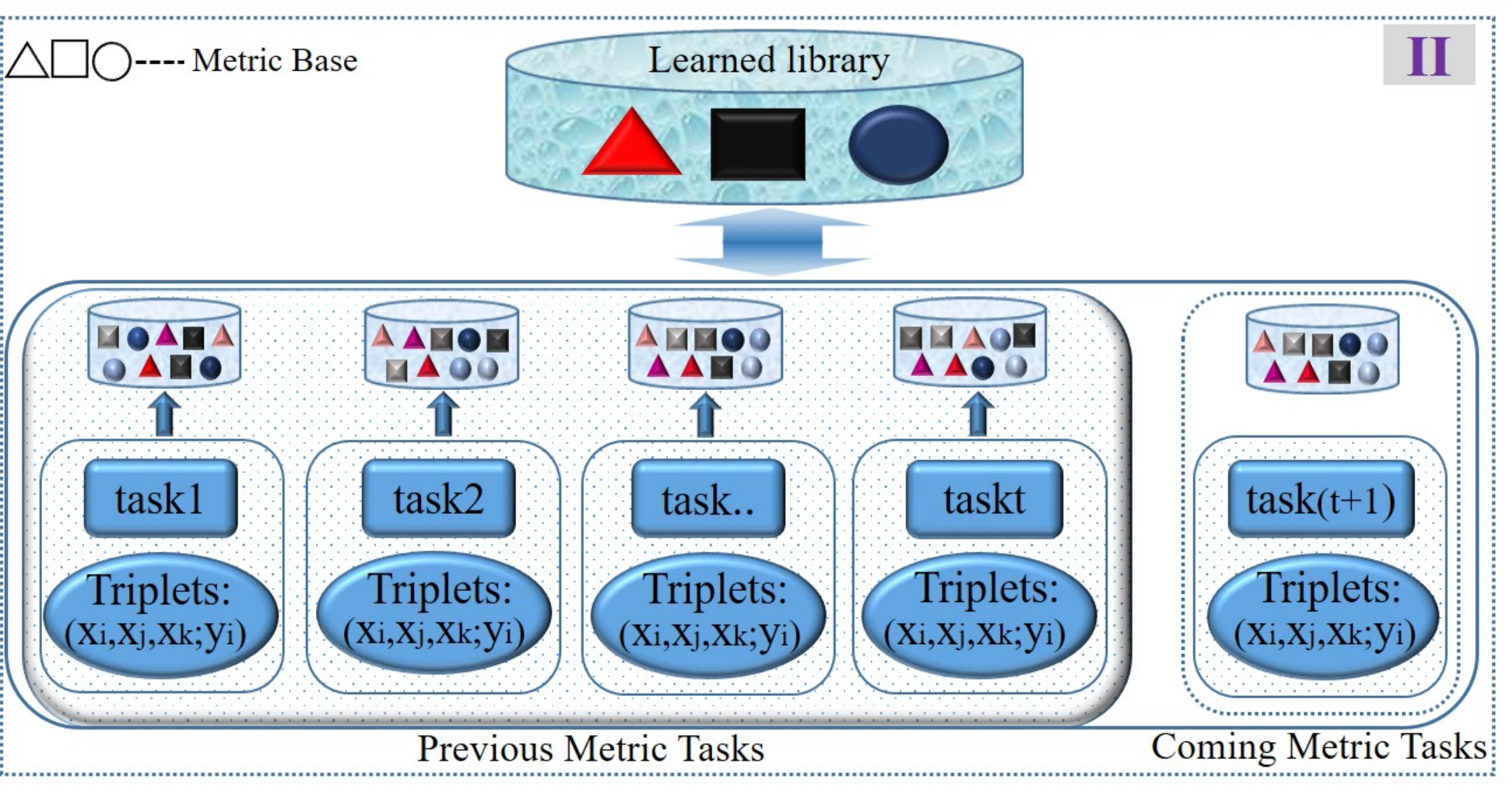}
        \end{minipage}}
       \vspace{-0.0mm}
    \caption{ The demonstration of the difference between lifelong metric learning and traditional multi-task metric learning: I) Our Lifelong Metric Learning. Knowledge in the learned library will be transferred to each new $t+1$-th metric task, and knowledge in the $t+1$-th task will be used to update the lifelong dictionary (metric base); II) Traditional metric learning utilizes all training data to update the knowledge in the library. Different shapes and colors denote different metric base (lifelong dictionary) and weights, respectively.}
  \label{fig:lifelongmetric}
  \vspace{-0pt}
 \end{figure*}

As depicted in Fig.~\ref{fig:lifelongmetric}, in this paper, we propose a new framework, called lifelong metric learning (LML), which intends to learn shared metric parameters from old ones without degrading performance or accessing to the old training data of $t$ tasks. Based on the assumption that all tasks are retained in a low-dimensional common subspace, LML learns a library called ``\textbf{lifelong dictionary}'' as a set of shared basis for all metric models, while the learned model of $t$ tasks can be considered as a sparse combination of this discriminative lifelong dictionary. Specifically, the lifelong dictionary can be initialized by extracting efficiently from the first training task at different regions via clustering. As new $t+1$-th task arrives, LML transfers knowledge through the shared base of lifelong dictionary to learn the new metric model with sparsity regularization, and refines the lifelong dictionary with first-order information from both the new task and previous tasks. By updating the lifelong dictionary continuously, the fresh knowledge is incorporated into the existing lifelong dictionary, thereby improving the performance of previously learned $t$ models. Therefore, model of new $t+1$-th task can be obtained without accessing to previous training data. To this end, we evaluate LML framework against state-of-the-art multi-task metric learning methods on several datasets. The experimental results validate encouraging performances of the proposed LML framework.

The contributions of this paper include:
\begin{itemize}
\item To the best of our knowledge, this is the first work about online metric learning from the perspective of lifelong learning, which adopts previous experience and knowledge of $t$ tasks to incorporate and learn the new $t+1$-th task, and can improve the performance in classification accuracy and reduce training time accordingly.
\item With the support of discriminative ``\textbf{lifelong dictionary}'', our proposed lifelong metric learning framework can model a new task via sparse combination, which can reduce the storage burden without saving the training data of previous $t$ tasks but first-order information.
\item We conduct comparisons and experiments with several real-world datasets, which verify the lower computational cost and higher improvement created by our LML framework.
\end{itemize}
The rest of the paper is structured as follows. Section \ref{sec:related work} gives a brief review of some related works. Section \ref{sec:formulation} introduces our proposed lifelong metric learning formulation. Section \ref{sec:convex} then proposes how to solve the proposed model efficiently via online passive aggressive optimization algorithm. In Section \ref{sec:experiment}, we report the experimental results and conclude this paper in Section \ref{sec:conclusion}.

\section{RELATED WORKS}\label{sec:related work}
Metric learning and its related methods have a long history. Depending on whether metric learning incorporates multi-task learning, metric learning can be roughly categorized as: \textbf{Single Metric Learning} and \textbf{Multi-task Metric Learning}.

\subsection{Single Metric Learning}
To the best of our knowledge, seeking a better distance metric through learning with a training dataset is at the key issue of of most state-of-the-art single metric learning models \cite{moutafis2017overview,yu2017deep,tao2015person}. For the distance metric based researches, the representative approaches can be categorized into two key issues: \textbf{batch metric learning} and \textbf{online metric learning}.

\indent The \textbf{batch metric learning} models \cite{mensink2013distance,niu2014information,hoi2010semi,wu2009learning,liu2015low,ding2017robust} can further be divided into two categories: models based on nearest neighbors, such as \cite{jacobgoldberger2004neighbourhood} optimizes the expected leave-one-out error of a stochastic nearest classifier in the projection space and \cite{weinberger2009distance} proposes the most widely-used Mahalanohis distance learning Large Margin Nearest Neighbors (LMNN), i.e., learning a Mahalanobis distance metric for $k$NN classification for labeled training examples; models based on pairs/triplets, for instance, \cite{xing2002distance} searches for a clustering that puts the similar pairs into the same clusters and dissimilar pairs into different clusters; \cite{lim2013robust} promotes input sparsity by imposing a group sparsity penalty on the learned metric and a trace constraint to encourage output sparsity; \cite{liu2015low} proposes a novel low-rank metric learning algorithm to yield bilinear similarity functions which can be applicable to high-dimensional data domains. However, \textbf{batch metric learning} models which assume all training samples are available prior to the learning phase cannot be applied into many practical applications, due to the fact that only a small amount of training samples are available in the beginning and others would come sequentially. Therefore, researchers focus on the online metric learning and intend to train the classifier with the new coming data.

For the \textbf{online metric learning}, \cite{chechik2009online} designs an Online Algorithm for Scalable Image Similarity learning (OASIS), for learning pairwise similarity that is fast and scales linearly with the number of objects and the number of non-zero features. However, OASIS may suffer from over-fitting and be difficult to be applied in the case of the high dimensions. Furthermore, computational complexity of learning full-rank metric can ranging from $O(d^2)$ to $O(d^{6.5})$, when metric learner lies in a high-dimensional sample space $\mb{R}^d$ and $d$ is the dimension of the training dataset. In order to overcome over-fitting problem, OMLLR \cite{cong2013self} proposes a novel online metric learning model with the low rank constraint, where low-rank metric enables to reduce storage of metric matrices. \cite{gao2014soml} incorporates large-scale high-dimensional dataset into sparse online metric learning, and explore its application to image retrieval. In addition, LORETA \cite{shalit2012online} describes an iterative online learning procedure, consisting of a gradient step, followed by a second-order retraction back to the manifold. To incorporate the benefits of both online learning and Mahalanobis distance, LEGO \cite{jain2009online} using a Log-Det regularization per instance loss, is guaranteed to yield a positive semidefinite matrix. Furthermore, more details can also be found in two surveys \cite{kulis2013metric} and \cite{bellet2013survey}.


\subsection{Multi-task Metric Learning}
Based on the assumption that the relationships and information shared among the different tasks can be taken into account, multi-task learning \cite{Caruana:1997,Chen:2011,Abhishek:2012,Obozinski:2009,Argyriou:2008,Jalali:2010,maurer2016benefit,yang2017latent} aims to improve generalization performance by learning multiple related tasks simultaneously. Furthermore, there are few multi-task metric learning methods designed to make metric learning benefit from training all tasks simultaneously. With the assumption that multiple tasks share a common Mahalanobis metric and each task has a task-specific metric, \textbf{mtLMNN} \cite{parameswaran2010large} adopts the LMNN formulation to the multi-task learning. However, \textbf{mtLMNN} is computationally more complicated, especially in the case of high dimensions. Specifically, there are $(t+1)d^2$ ($t$ and $d$ denote the task number and data dimension, respectively) parameters to be optimized. Based on low-rank based assumptions, \cite{yang2011multi} presents transformation matrix to the problem of multi-task metric learning by learning a common subspace for all tasks and an individual metric for each task, where each individual metric is restricted in the common subspace. In addition, \textbf{mtSCML} \cite{shi2014sparse} constructs a common basis set, multi-metric are regularized to be relevant across tasks (as favored by the group sparsity). However, storage and computation will become cumbersome with large scale tasks. Therefore, in order to address the situation that total number of tasks is large or the task is coming consecutively, we employ the common subspace as the lifelong dictionary, and then build a more robust lifelong metric learning framework.



\textbf{Notations:}
For matrix $W\in \mathbb{R}^{m\times n}$, let $w_{ij}$ be the entry in the $i$-th row and $j$-th column of $W$. Let us define some norms, $\left\|W\right\|_0$ is the number of nonzero entries in $W$; denote by $\left\|W\right\|_1=\sum_{i=1}^m\sum_{j=1}^n|w_{ij}|$ and $\left\|W\right\|_{\infty}=\max_{i,j}|w_{ij}|$ the $\ell_1$-norm and $\ell_{\infty}$-norm of $W$, respectively. Let $\left\|W\right\|_{2,1}=\sum_{i=1}^m\left\|w_i\right\|_2$; denote by $\mr{sgn}(W)$, $(W)_+$ and $\lvert \cdot\lvert$ the elementwise sign, positive part elementwise and absolute value of matrix $W$, respectively. Let $\odot$ be the elementwise multiplication.

\section{Lifelong Metric Learning }\label{sec:formulation}
\subsection{Preliminaries}
Assume that there are $m$ related tasks. $(X_t,Y_t)$ denotes the training set to the $t$-th task with $\{x_{ti}\in \mb{R}^{\hat{d}}, i=1,\ldots,n_{t}\}$, where $\hat{d}$ and $n_t$ are the dimension and the number of the training samples of $t$-th task, respectively. Define $n=\sum_{t=1}^m n_{t} $ to be the total number of samples, $m$ is the total number of tasks and $f_t: \mb{R}^{\hat{d}}\times \mb{R}^{\hat{d}} \rightarrow \mb{R}$ to be the similarity / distance metric of the $t$-th learning tasks. The $f_t$ is assumed to be defined based on a linear transformation $L_t:\mb{R}^{\hat{d}}\rightarrow \mb{R}^d$ (with $d\ll \hat{d}$ to obtain a low dimensional representation) as:
\begin{itemize}
  \item \textbf{Similarity Function:}
  \begin{equation}\label{eq:simi_l}
   f_{L_t}(x_{ti},x_{tj})=x_{ti}^TL_{t}^TL_{t}x_{tj}=: f_{t,ij}(L_{t}^TL_{t}).
  \end{equation}
  \item \textbf{Distance Function:}
  \begin{equation}\label{eq:dist_l}
  f_{L_t}(x_{ti},x_{tj})=\triangle x_{t,ij}^TL_{t}^TL_{t}\triangle x_{t,ij}=: f_{t,ij}(L_{t}^TL_{t}),
  \end{equation}
\end{itemize}
where $x_{ti}$ and $x_{tj}$ are feature vectors, and $\triangle x_{t,ij}=x_{ti}-x_{tj}$. $L_t^TL_t \in \mb{R}^{\hat{d}\times \hat{d}}$ must be positive semi-definite to satisfy the properties of a similarity / distance metric. The set of triplets $\mc{T}_t=\{(i,j,k)|(i,j)\in \mc{S}_t,(i,k)\in \mc{D}_t\}$ are used to define the side-information in $X_t$, where $\mc{S}_{t}$ and $\mc{D}_{t}$ denote all the similar and dissimilar pairs, respectively. For example, $f_{L_t}(x_{ti},x_{tj})\leq f_{L_t}(x_{ti},x_{tk})$  implies similar data pairs $\{(x_{ti},x_{tj})|(i,j)\in \mc{S}_t\}$ to stay closer than dissimilar pairs $\{(x_{ti},x_{tk})|(i,k)\in \mc{D}_t\}$ depending on the similarity / distance metric $f_t$. Without specially specifying, the similarity and distance function are denoted as $f_{L_t}(x_{ti},x_{tj})$ in the following.

\subsection{The Lifelong Metric Learning Problem}
The original intention of multi-task metric learning is to learn an appropriate distance metric $f_t$ for $t$-th task  utilizing all the side-information from the joint training set $\{(X_1,Y_1),(X_2,Y_2),\ldots, (X_m,Y_m)\}$. Suppose that the loss involved in $t$-th task is determined by the distance function $f_t$ (with metric $L_t$) and the pairs appearing in $\mc{S}_t$ and $\mc{D}_t$:
\begin{equation}\label{eq:basic_ell}
\ell_{t}(L_{t}^TL_{t})=\ell_{t}\Big(f_{t,ij}(L_{t}^TL_{t})\Big), \quad \forall (i,j)\in \mc{S}_t\cup \mc{D}_t,
\end{equation}
where $\ell_{t}$ is an arbitrary loss function of $t$-th task. However, \textbf{learning new metric task without accessing to the previously used training data} is not considered by traditional multi-task metric learning.
In the context of multi-task metric learning, a lifelong metric learning system encounters a series of metric learning tasks $\ell_1,\ell_2,\ldots,\ell_m$, where each task $\ell_t$ is defined by Eq. \eqref{eq:basic_ell}. For convenience, we do not assume that the learner knows any information about tasks, e.g., the total number of tasks $m$, the distribution of these tasks, etc. In each time step, as the lifelong system receives a batch of training data for some metric learning task $t$, either a new metric task or previously learning task, this system may be asked to make predictions on samples of any previous task. Its goal is to establish task models $L_1,\ldots,L_m$ such that:
\begin{itemize}
  \item Classification Accuracy: each learned metric $L_t$ should classify the new samples more accurate. 
  \item Computation Efficiency: in the training period, each $L_t$ should be updated faster than traditional multi-task metric learning (i.e., joint learning models). 
  \item Lifelong Learning: new $L_t$'s can be added arbitrarily and efficiently when the lifelong system encounters new metric tasks.
\end{itemize}


\begin{figure}[t]
\centering
\includegraphics[width =245pt,height =73pt]{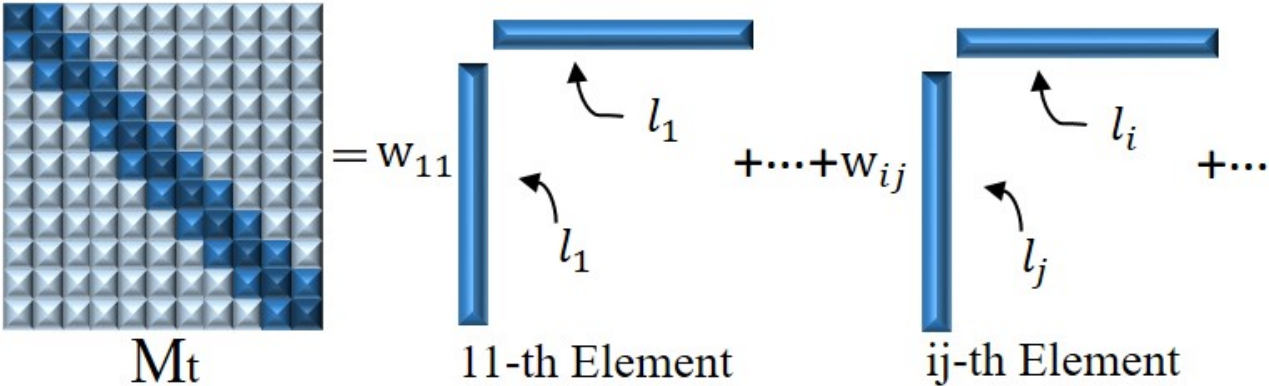}
\vspace{-2.0pt}
\caption{The demonstration of formulation given by Eq.~\eqref{eq:our_decomposition}, where $t$-th task can be represented by a series of ``atoms'' in the lifelong dictionary, and $\mr{w}_{11}$ and $\mr{w}_{ij}$ are corresponding weights of the $11$-th element and $ij$-th element, respectively.}
\label{fig:decomposition}
 \vspace{-1pt}
\end{figure}

\subsection{Lifelong Metric Learning Framework (LML)}
In order to model the correlation among different metric tasks, we assume that the metric matrix $f_t$ for $t$-th task can be represented using a combination of the shared common subspace from a knowledge repository. Moreover, motivated by \cite{yang2011multi},  \emph{Theorem 1} gives the detail mathematical description.
\begin{theorem}
Let $f_{L_t}(x_{ti},x_{tj})$ denotes the similarity / distance of $x_{ti},x_{tj}\in \mb{R}^{\hat{d}}$ defined by the transformation matrix $L_t$ as Eq. \eqref{eq:simi_l} or Eq. \eqref{eq:dist_l}. For any $L_t\in\mb{R}^{d\times \hat{d}}$ ($d\ll \hat{d}$), there exists a low dimensional subspace $\mb{S}_t$ spanned by orthonormal basis $\{p_{t1},\ldots,p_{td}\}$ with metric matrix defined by $R_t\in \mb{R}^{d\times d}$ so that
\[f_{L_t}(x_{ti},x_{tj})=f_{R_{t}}(\hat{x}_{ti},\hat{x}_{tj}),\]
where $\hat{x}_{ti}=P_t^Tx_{ti}=[p_{ti},\ldots,p_{td}]^Tx_{ti}\in \mb{R}^d$ is the coordinate of the projection of $x_{ti}$ in $\mb{S}_t$ with respect to basis matrix $P_{t}$.
\end{theorem}

Therefore, metric matrix $L_t$ for $t$-th task in \emph{Theorem 1} can be explicitly decomposed to a low-dimensional metric part $R_t$ and  a subspace part $P_t$.  Our Lifelong Metric Learning (LML) framework can be simply represented as to learn an individual metric $R_t$ for each task in a common subspace $P_t^T=L_0$. Furthermore, as shown in Fig.~\ref{fig:decomposition}, parameter matrix $M_t\in \mb{R}^{\hat{d}\times \hat{d}}$ for metric task $f_t$ can be expressed as:
\begin{equation}\label{eq:our_decomposition}
M_t=L_t^TL_t=L_0^TR_t^TR_tL_0=L_0^TW_tL_0=\sum_{i=1}^d\sum_{j=1}^d w_{ij}l_il_j,
\end{equation}
where $W_t\in \mb{R}^{d\times d}$ denotes the weight matrix. Therefore,  each metric task $M_t$ can be represented as a linear combination of ``\textbf{lifelong dictionary}'' composed by $l_il_j, \forall i,j=1,\ldots, d $. Generally, since diagonal elements in $W_t$ represents the self-correlation of a transformed feature while off-diagonal element represents correlation among different transformed features, diagonal elements should be more dense than those off-diagonal elements. We encourage the off-diagonal elements of $W_t$'s to be sparse (i.e., use few components among lifelong dictionary) in order to ensure that each learned metric model captures a maximal reusable chunk of knowledge.

Given the training data for each task, we optimize the metrics to minimize the loss function over all tasks while encouraging the metrics to share common knowledge in lifelong dictionary. Therefore, LML framework can be formulated as:
\begin{equation}\label{eq:model1}
\begin{aligned}
\min_{L_0,\{W_t\}} &\frac{1}{m}\Big\{\sum_{t=1}^m \ell_{t}(L_0^TW_tL_0) \Big)+\lambda_t\left\|W_t\right\|_{\mr{1,off}}\Big\}  \\
&+\gamma \left\|L_0\right\|_F^2,
\end{aligned}
\end{equation}
where the $\left\|\cdot\right\|_{\mr{1,off}}$-norm of $W_t$ defined as $\sum_{i \neq j}|W_{t,ij}|$ is used as a convex approximation to the true matrix sparsity, and $\left\|L_0\right\|_F=(\mr{tr}(L_0L_0^T))^{1/2}$ is the Frobenius norm of matrix $L_0$ to avoid overfitting. The trade-off parameter $\lambda_t\geq 0$ controls the regularization of $\left\|W_t\right\|_{\mr{1,off}}$ for all $t=1,\ldots,m$. If $\lambda_t\rightarrow \infty$, the task-specific matrices $W_t$'s become self-correlation diagonal matrices. With the definition of $\ell_t$ in Eq.~\eqref{eq:basic_ell}, the final optimization problem of lifelong metric learning can be formulated as:
\begin{equation}\label{eq:model_final}
\begin{aligned}
\min_{L_0,\{W_t\}} &\frac{1}{m}\Big\{\sum_{t=1}^m \ell_{t}\Big(f_{t,ij}(L_0^TW_tL_0) \Big)+\lambda_t\left\|W_t\right\|_{\mr{1,off}}\Big\}  \\
&+\gamma \left\|L_0\right\|_F^2,
\end{aligned}
\end{equation}
where $(i,j)\in \mc{S}_t\cup\mc{D}_t$ are the side-information in the $t$-th metric task.

\section{Model Optimization}\label{sec:convex}
This section provides the detail procedure of how to optimize our proposed LML framework. Since the problem in Eq.~\eqref{eq:model_final} is not convex with respect to $L_0$ and $W_t$'s jointly, the objective function can arrive at a local optimum. A common approach for computing such a local optimum for objective functions in Eq.~\eqref{eq:model_final} is to alternately perform two convex optimization steps: one in which $L_0$ is optimized by fixing the $W_t$'s, and another in which the $W_t$'s are optimized by holding $L_0$ fixed. However, as shown in \cite{ruvolo2013ella}, this approach is inefficient and inapplicable to lifelong learning with many tasks and data samples. This is because that in order to optimize $L_0$, the problem in Eq.~\eqref{eq:model_final} has to recompute the value of each $W_t$'s (which will become time consumption when increasing the number of learned tasks $m$). To address this problem, we aim to approximate Eq.~\eqref{eq:model_final} by applying the online passive aggressive (PA) \cite{crammer2006online} optimization strategy, i.e.,
\begin{equation}\label{eq:model_pa}
\begin{aligned}
\min_{L_0,\{W_t\}} & \frac{1}{m}\Big\{\sum_{t=1}^m \ell_{t}\Big( f_{t,ij}(L_0^TW_tL_0) \Big)+\frac{1}{2\eta}\left\|L_0^TW_tL_0-M_t\right\|_F^2 \\ &+\lambda_t\left\|W_t\right\|_{1,\mr{off}} \Big \}+\gamma \left\|L_0\right\|_F^2,
\end{aligned}
\end{equation}
where $\eta$ is the learning rate. After linearizing the loss function $\ell_{t}$ around $L_0^TW_tL_0=M_t$, we obtain the following new online function:
\begin{equation}\label{eq:model_line}
\begin{aligned}
& \min_{L_0,\{W_t\}}  \frac{1}{m}\Big\{\sum_{t=1}^m \ell_{t}\Big( f_{t,ij}(M_t) \Big)+\langle L_0^TW_tL_0-M_t, G_t \rangle \\
& +\frac{1}{2\eta}\left\|L_0^TW_tL_0-M_t\right\|_F^2+\lambda_t\left\|W_t\right\|_{1,\mr{off}}\Big\}+\gamma \left\|L_0\right\|_F^2,
\end{aligned}
\end{equation}
where $G_t$ is the gradient of $\ell_{t}$. We then rewrite the optimization problem in Eq.~\eqref{eq:model_line} as:
\begin{equation}\label{eq:model_line2}
\begin{aligned}
\min_{L_0,\{W_t\}} & \frac{1}{m}\Big\{\sum_{t=1}^m    \left\|L_0^TW_tL_0-(M_t-\eta G_t)\right\|_F^2+\lambda_t\left\|W_t\right\|_{1,\mr{off}} \Big\}  \\
& +\gamma \left\|L_0\right\|_F^2.
\end{aligned}
\end{equation}
 In Eq.~\eqref{eq:model_line2}, we have suppressed the constant term of the linearize form (since it does not affect the minimum). Crucially, we have removed the dependence of the optimization problem Eq.~\eqref{eq:model_final} on the number of the data samples $n_1,\ldots,n_t$ in each task. Additionally, Eq.~\eqref{eq:model_line2} can be reformulate as:
\begin{equation}\label{eq:model_line3}
\begin{aligned}
\min_{L_0,\{W_t\}} & \frac{1}{m}\Big\{\sum_{t=1}^m  \left\|L_0^TW_tL_0-M_t^*\right\|_F^2+\lambda_t\left\|W_t\right\|_{1,\mr{off}} \Big\}  \\
&+\gamma \left\|L_0\right\|_F^2,
\end{aligned}
\end{equation}
where $M_t^*=M_t-\eta G_t$ can be approximated from the large samples by online learning or small samples by offline mini-batch learning in the $t$-th task. Moreover, the optimization problem in Eq.~\eqref{eq:model_line3} also can be roughly divided into two subproblems with alternating direction optimization strategy. After initializing the lifelong dictionary $L_0$, the first subproblem is to compute the optimal $W_t$ for the new coming task $M_t^*$, and the second subproblem is to update the lifelong dictionary $L_0$ by fixing $W_t$'s.

\renewcommand{\algorithmicrequire}{\textbf{Input:}}
\renewcommand{\algorithmicensure}{\textbf{Output:}}
\begin{algorithm}[t]
\caption{\small Proximal Method for Solving $W_t$}
\begin{algorithmic}[1]
\REQUIRE $W_t^0$, $V_0\in \mathbb{R}^{d\times d}$, $\lambda_t\geq 0,\eta_0\geq 0,$ and MAX-ITER
\ENSURE $W_t$ \\
\STATE Initialize $W_t^1=W_t^0$, $t_{-1}=0$, $t_0=1$
\FOR {$i=1,...,$ MAX-ITER}
\STATE  $V_{i-1}=W_t^i+\alpha_i(W_t^i-W_t^{i-1})$;
\WHILE {true}
\STATE Compute $V_i$ via Eq. \eqref{eq:solution_w};
\STATE Update $\eta_i$ via backtracking rule.
\ENDWHILE
\STATE $W_t^{i+1}=V_i$, $\eta_{i+1}=\eta_i$;
\IF   {Convergence criteria satisfied}
\STATE   $W_t=W_t^{i+1}$;
\STATE break;
\ENDIF
\ENDFOR \\
\STATE Return $W_t$;
\end{algorithmic}
\end{algorithm}
\vspace{1.0mm}


\renewcommand{\algorithmicrequire}{\textbf{Input:}}
\renewcommand{\algorithmicensure}{\textbf{Output:}}
\begin{algorithm}[t]
\caption{\small Lifelong Metric Learning Framework }
\begin{algorithmic}[1]
\REQUIRE Training dataset $(X_1,Y_1), \ldots, (X_t,Y_t)$, $\hat{d}\geq d>0,\{\lambda_t\},\gamma$;
\ENSURE $\{W_t\}, L_0$ \\
\STATE Initialize $L_0$ in the first coming task;
\WHILE {Not Converge}
\STATE  New $t$-th task: $(X_{\mr{new}},Y_{\mr{new}},t)$
\IF {isNewTask($t$) }
\STATE $T \leftarrow T+1$
\STATE $X_{t}\leftarrow X_{\mr{new}}$, $Y_{t}\leftarrow Y_{\mr{new}}$
\ELSE
\STATE $X_t\leftarrow [ X_t, X_{\mr{new}}]$, $Y_t\leftarrow [Y_t,Y_{\mr{new}}]$
\ENDIF
\STATE $M_t\leftarrow \mr{SingleTaskLearner (X_t, Y_t)}$;
\STATE  Update $W_t$ via \textbf{Algorithm 1};
\STATE  Update $L_0$ via Gradient Method;
\ENDWHILE \\
\STATE Return $W_t$;
\end{algorithmic}
\end{algorithm}

\subsection{Lifelong Dictionary $L_0$ Initialization}
An high-quality lifelong dictionary plays an important role in our model. In order to generate a set of discriminative basis vectors in $L_0$, we first divide data into different clusters. For each clutter, we select $J$ nearest neighbors from each class (for $J=|{10,20,50|}$ to count for different scales), and apply Fisher discriminative analysis followed by eigenvalue decomposition to obtain the basis elements.

\subsection{Solve $W_t$ with Given $L_0$}
 With the initialized lifelong dictionary $L_0$, $W_t$ is the variable in this subproblem. The optimization function can be rewritten as:
\begin{equation}\label{eq:model_samplity}
\begin{aligned}
\min_{W_t}   f(W_t)+g(W_t),
\end{aligned}
\end{equation}
where $f(W_t)=\frac{1}{2}\left\|L_0^TW_tL_0-M_t^*\right\|_F^2$ and $g(W_t)=\lambda_t\left\|W_t\right\|_{1,\mr{off}}$. Due to the non-smooth nature of $g(W_t)$, we propose the proximal gradient method (FISTA) \cite{beck2009fast} with a fast global convergence rate to solve this optimization problem. Specifically, the proximal operator of  the $\ell_{1,\mr{off}}$-norm can be applied to solve this subproblem:
\begin{equation}\label{eq:solution_w}
W_t^{i+1}=\arg\min_{W} \frac{1}{2}\left\|W-W_t^i+\eta_i \nabla f(W_t^i)\right\|_F^2+\lambda_t\left\|W\right\|_{1,\mr{off}},
\end{equation}
where $\eta_i>0$ is the stepsize parameter, Eq.~\eqref{eq:solution_w} can be appropriately determined by the backtracking rule. $\nabla f(W_t^i)$ is the gradient matrix with respect to $ f(W_t^i)$ can be expressed as:
\[\nabla f(W_t^i)=L_0L_0^TW_t^iL_0L_0^T-L_0M_tL_0^T.\]
With the gradient of $ f(W_t^i)$, the optimal $W_t^{i+1}$ depends on the proximity operator of the $\ell_{1,\mr{off}}$-norm, i.e., soft thresholding operator:
\begin{equation}\label{ensure w}
 \mr{pros}(W,\lambda_t\eta_i)=\mr{sgn}(W)\odot\Big(\lvert W\lvert-\lambda_t\eta_i(1-I)\Big)_+,
\end{equation}
where $\odot$ denotes the elementwise multiplication. Notice that FISTA amounts for using two sequences $\{W_t^i\}$ and $\{V_t^i\}$ in which $\{W_t^i\}$ is the approximate solution and $\{V_t^i\}$ is search points. Moreover, the proximal method by \textbf{Solving for $W_t$} is summarized in \textbf{Algorithm 1}.


\begin{table*}[t]
\caption{Statistics of the Base Metric Learning Models}
\centering
\scalebox{0.96}{
\begin{tabular}{c|c|c|c}
\hline
{Name}&Metric Type&Metric Function & $\triangle_t$ in Eq.~\eqref{eq:gradient_l0} \\
 \hline
  \centering \textbf{OASIS} \cite{chechik2009online} & Similarity &$\mr{\ell_t(M_t)=\mr{max}\big(0,1-s_{M_t}(x_i,x_j)+s_{M_t}(x_i,x_k)\big)}$  &$\mr{ \sum_{(i,j,k)\in \mc{T}}\big(x_i(x_k-x_j)^T+(x_k-x_j)x_i^T \big) }$\\
   \hline
   \centering \textbf{SCML} \cite{shi2014sparse} &Distance & $\mr{\ell_t(M_t)=\mr{max}\big(0,1-d_{M_t}(x_i,x_j)+d_{M_t}(x_i,x_k)\big) }$ & $\mr{ \sum_{(i,j,k)\in \mc{T}} \Big( (x_i-x_j)(x_i-x_j)^T-(x_i-x_k)(x_i-x_k)^T \Big)}$ \\
  \hline
\end{tabular}
}
\label{table:base_model}
\end{table*}

\subsection{Solve $L_0$ with Given $\mc{T}_t$'s and $W_t$'s}
In order to evaluate the lifelong dictionary $L_0$, we modify the formulation in Eq.~\eqref{eq:model_pa} to remove the minimization over $W_t$. Besides, we also remove the second term which is used to keep the new similarity / distance matrix close to the current one. Further, we accomplish this by exploiting both side-information $\mc{T}_t$ (generated according to the adopted base metric learning model) and $W_t$ in the learned tasks. In the following, we try to adopt the gradient descent method to solve $L_0$ in Eq.~\eqref{eq:model_pa}. The gradient of $\ell_t$ with respect to $L_0$ is:
\begin{equation}
\begin{small}
\begin{aligned}\label{eq:gradient_l0}
\frac{1}{m}\sum_{t=1}^m & \frac{\partial\ell_{t}\Big( f_{t,ij}(L_0^TW_tL_0)\Big)}{\partial L_0}+ \frac{\partial (\gamma \left\|L_0\right\|_F^2)}{\partial L_0}, \\
=\frac{1}{m}\sum_{t=1}^m & \big(W_t^TL_0\triangle_t\big)+\gamma L_0,
\end{aligned}
\vspace{-2.0pt}
\end{small}
\end{equation}
where $\triangle_t$ can be calculated with different SingleTaskLearner function using side-information $\mc{T}_t$. The LML framework is summarized in \textbf{Algorithm 2}, where SingleTaskLearner is learned using base metric models.
\vspace{-2.0pt}

\subsection{Computational Complexity}
For the complexity of our proposed algorithm, the main computational cost in each update in \textbf{Algorithm 2} involves two subproblems: one optimization problem lies in Eq.~\eqref{eq:model_samplity}, another one is Eq.~\eqref{eq:gradient_l0}.
\begin{enumerate}
  \item For the problem in Eq.~\eqref{eq:model_samplity}, each update for the LML system begins by base metric learning models to compute $M_t$, we assume that this step has complexity $O(\xi (\hat{d},n))$, where $\hat{d}$ is the number of feature, $n=\sum_{t=1}^mn_t$ and $n_t$ is the triplets number of $\mc{T}_t$ in our paper. Next, to update $W_t$ requires solving the instance of lasso, i.e., $\left||W\right\|_{1,\mr{off}}$.  Each iteration in this problem begins by the computation of the gradient of $W_t$, and the computational complexity is $O(d^2\hat{d}+2d^3+2\hat{d}^3)$. Therefore, the cost for achieving $\epsilon$-accuracy is $O((d^2\hat{d}+2d^3+2\hat{d}^3+\xi (\hat{d},n))/\sqrt{\epsilon})$, where $\epsilon$ is determined by the convergence property of Accelerated Gradient Method, i.e., $\mc{O}(1/\sqrt{\epsilon})$ with
      \begin{equation}\label{eq:optimal_solution_s}
     \begin{aligned}
      & f(W_t)+g(W_t)-f(W^*)-g(W^*)  \\
      &\leq \sqrt{\frac{2\eta_t\left\|W_0-W^*\right\|_F^2}{\epsilon}}-1.
      \end{aligned}
      \end{equation}
      In other words, the convergence rate of \textbf{Algorithm 1} can achieve $\mc{O}(1/T^2)$ as shown in \cite{beck2009fast}. Moreover, the multiplication of two matrices can be further reduced with Coppersmith-Winograd algorithm.

  \item The optimization algorithm for solving $L_0$ involves the gradient of each triplet in $\mc{T}_t$, and the computational complexity is $O(\hat{d}^2d+\hat{d}d^2)$.
\end{enumerate}
Finally, the overall complexity of each update in \textbf{Algorithm 2} is $O(\hat{d}^2d+\hat{d}d^2+(d^2\hat{d}+2d^3+2\hat{d}^3+\xi (\hat{d},n))/\sqrt{\epsilon})$.

\subsection{Discussion}
In this section, we briefly review one learning method that is most related to our proposed learning algorithm. Perhaps the most relevant work to ours in the context of multi-task metric learning is from \cite{shi2014sparse}, which frames metric learning as learning a sparse combination of locally discriminative metrics that are generated from the training data via clustering. However, the motivation for \textbf{SCML} and our LML are significantly different:
\begin{itemize}
  \item \textbf{SCML} aims to cast metric learning as learning a \emph{sparse combination of basis elements} taken from a basis set $B=\{b_i\}_{i=1}^K$, where the $b_i$'s are $\hat{d}$-dimensional column vectors. Instead of fixing the metric task number, our LML focuses on the transfer of knowledge from preciously learned tasks to the new metric task using the shared basis, i.e., lifelong task learning.
  \item In \textbf{SCML}, the metric matrix is represented using basis matrices induced by a $\ell_1$-norm constraint, and the formulation in \textbf{SCML} can only achieve batch learning. Our LML encourages the communication among different \emph{basis elements} via a $\ell_{1,\mr{off}}$-norm constraint, and the resulting formulation can integrate online sample learning by adopting the SinlgeTaskLearner as online metric learning. Furthermore, we have also conducted extensive experiments on the effect of $\ell_{1,\mr{off}}$-norm in the experiment section.
  \item The optimization algorithm in \textbf{SCML} can only find a local solution with all the training samples. The proposed algorithm for Eq.~\eqref{eq:model_line3} can learn new metric task without accessing to historical data because only the gradient information of previous training data is adopted in the next iteration in Eq.~\eqref{eq:gradient_l0}.
\end{itemize}


\section{Experiments}\label{sec:experiment}
In this section, we carry out empirical comparisons with the state-of-the-art single and multi-task metric learning models. We first give the base metric learning with our lifelong metric learning framework in Table \ref{table:base_model} with two different function: ${s_{M_t}(x_i,x_j)=x_i^TM_tx_j}$ and ${d_{M_t}(x_i,x_j)=(x_i-x_j)^TM_t(x_i-x_j)}$, where  $x_i$ and $x_j$ belong to the same class, while $x_i$ and $x_k$ are from different classes. The experiments are then conducted on a series of real datasets.


\subsection{Comparison Algorithms and Evaluation}
In our experiments, we compare our LML framework with single metric learning models and multi-task metric learning models. The single metric learning model includes:
1) Euclidean distance (\textbf{stEuc}): the standard Euclidean distance in feature space; 2) OASIS (\textbf{stOASIS}) \cite{chechik2009online}: the classical online metric learning model which is given in Table~\ref{table:base_model}, and its iteration number is $2\times 10^{4}$ in our paper; 3)
LMNN (\textbf{stLMNN}) \cite{weinberger2009distance}: Large Margin Nearest Neighbor Classification, which learns a Mahalanobis distance for $k$-nearest neighbor classification; 4) SCML-global (\textbf{stSCML}) \cite{shi2014sparse}: which is simply to combine the local basis elements into a higher-rank global metric; 5) LMNN-union (\textbf{uLMNN}): is the LMNN metric obtained on the union of the training data of all tasks (i.e., ``pooling'' all the training data and ignoring the multi-task aspect). 6) SCML-union (\textbf{uSCML}): is the SCML metric obtained on the union of the training data of all metric tasks.

For the multi-task metric learning models, the comparison models include:
\begin{itemize}
  \item multi-task LMNN (\textbf{mtLMNN}) \cite{parameswaran2010large}: common metric defined by $M_0$ picks up general trends across multiple datasets and $M_t$ specializes the metric further for each particular task.
  \item multi-task SCML (\textbf{mtSCML})  \cite{shi2014sparse}: this multi-task metric learning model considers that all learned metrics can be expressed as combinations of the same basis subset $B$, though with different weights for each task.
\end{itemize}

 For the classical lifelong multi-task learning, we adopt the comparison model as:
 \begin{itemize}
   \item Lifelong multi-task (\textbf{ELLA}) \cite{ruvolo2013ella}: whose formulation is realized by the following objective function:
       \begin{equation}
        \begin{aligned}
        \frac{1}{m}\sum_{t=1}^m \min_{s^{(t)}} & \Big\{\frac{1}{n_t}\sum_{i=1}^{n_t}\mc{L}\Big(f(x_i^{(t)};Ls^{(t)}), y_i^{(t)} \Big)+\mu\left\|s^{(t)}\right\|_1 \Big\} \\
         &+\lambda \left\|L\right\|_F^2,
    \end{aligned}
    \end{equation}
 where $(x_i^{(t)},y_i^{(t)})$ is the $i$-th labeled training samples for $t$-th task, $\mc{L}$ is a known loss function. Specifically, \textbf{ELLA} maintains a sparsely shared basis vector for all regression or logistic task models, transfers knowledge from the basis to learn new $t$-th task.
 \end{itemize}

 All the models are implemented in MATLAB, and the codes are available at the supplement website. Notice that all the parameters of the models are tuned in $\{10, 1, 0.1, 0.01 ,0.001\}$ and selected via 5-fold cross validation. Although our model allows different weights $\lambda_t$ for each task, throughout this paper we only adjust our parameters: $\gamma$ and $\lambda=\lambda_{t}>0 $. All the experiments are performed on the computer with 12G RAM, Intel i7 CPU.


\begin{table*}[t]
\caption{Sentiment $\&$ Isolet dataset: classification error and training time of the competing metric learning models. The reported performance is averaged over five random repetitions, and methods with the best performance are marked as bolded black.}
\scriptsize
\centering
\scalebox{0.94}{
\begin{tabular}{p{0.83cm}|c||c|c|c|c||c|c||c|c||c|c|c}
\hline
{Dataset}&Task &stEuc &stLMNN  & stSCML & stOASIS &uLMNN & uSCML  & mtLMNN & mtSCML &ELLA & Ours+OASIS& Ours+SCML \\
 \hline
 \hline
   \multirow {7}{*}{
  \centering \textbf{\Large {$\mr{Senti}\atop \mr{ment}$} }}
   &\textbf{Books}       & 33.5 $\pm$0.5&29.7$\pm$0.4& 27.0$\pm$0.5& 28.3$\pm$0.4 & 29.6$\pm$0.4& 28.0$\pm$0.4& 29.1$\pm$0.4& 25.8$\pm$0.4& 32.8$\pm$0.5& 27.8$\pm$0.4 &\textbf{25.3 $\pm$0.5 }\\
  &\textbf{DVD}          & 33.9$\pm$0.5&29.4$\pm$0.5& 26.8$\pm$0.4& 23.5$\pm$0.4 & 29.4$\pm$0.5& 27.9$\pm$0.5& 29.5$\pm$0.5& 26.5$\pm$0.5& 31.0$\pm$0.7& \textbf{23.5$\pm$0.5}& 25.0$\pm$0.4  \\
   &\textbf{Electronics} & 26.2$\pm$0.4 &23.3$\pm$0.4& 21.1$\pm$0.5& 20.3$\pm$0.4 & 25.1$\pm$0.4& 22.9$\pm$0.4& 22.5$\pm$0.4& 20.2$\pm$0.5& 19.0$\pm$0.7& \textbf{18.0$\pm$0.4}& 18.5$\pm$0.4 \\
    & \textbf{Kitchen}   & 26.2$\pm$0.6&21.2$\pm$0.5& 19.0$\pm$0.4& 17.3$\pm$0.4 & 23.5$\pm$0.3& 21.9$\pm$0.5&22.1$\pm$0.5&19.0$\pm$0.4& 16.1 $\pm$0.5& \textbf{12.0$\pm$0.4} &15.8$\pm$0.4 \\
    &\textbf{Avg. Error}  &30.0$\pm$0.2&25.9$\pm$0.2& 23.5$\pm$0.2& 22.4$\pm$0.3 & 26.9$\pm$0.2 &25.2$\pm$0.2 &25.8$\pm$0.2 & 22.9$\pm$0.2& 24.8 $\pm$ 0.4 & \textbf{20.3$\pm$0.2} &21.2 $\pm$0.4  \\
    &\textbf{Avg. Runtime} & N/A & 11min& 12s & 0.5min & 9min& 10s &8min&1min &0.5min & 0.5min & 18s  \\
    \hline
   \multirow {8}{*}{
  \centering \large Isolet}
     &\textbf{Isolet1}  & 28.9 $\pm$0.0&23.2$\pm$0.1& 19.6$\pm$0.2& 24.5$\pm$0.2 & 23.2$\pm$0.1& 55.3$\pm$0.1& 21.3$\pm$0.7&19.1$\pm$0.2&N/A&  21.7$\pm$ 0& \textbf{ 16.5$\pm$0.1 } \\
  &\textbf{Isolet2}  & 30.5$\pm$0.7&24.4$\pm$0.9& 20.9$\pm$1.6& 19.9$\pm$0.0 & 24.4$\pm$0.9& 52.9$\pm$1.0& 22.9$\pm$0.6& 20.1$\pm$2.1&N/A& \textbf{23.3$\pm$1.3}& 22.4$\pm$2.1 \\
  &\textbf{Isolet3}  & 35.3$\pm$1.2&28.4$\pm$1.2& 24.5$\pm$0.3& 25.8$\pm$0.8 & 28.4$\pm$1.2& 53.1$\pm$2.8& 26.0$\pm$1.2& 22.9$\pm$0.2&N/A& \textbf{21.0$\pm$1.4}& 23.3$\pm$0.0  \\
  &\textbf{Isolet4}  & 35.7$\pm$0.4&27.2$\pm$1.9& 25.3$\pm$2.4& 29.4$\pm$2.7 & 27.2$\pm$1.9& 53.5$\pm$1.8&25.3$\pm$2.4&23.8$\pm$2.8&N/A& \textbf{23.1$\pm$1.0} &25.1$\pm$0.2 \\
  &\textbf{Isolet5}  & 37.4$\pm$0.5&30.2$\pm$0.8& 26.7$\pm$1.0& 29.7$\pm$0.8 & 30.2$\pm$0.8& 54.6$\pm$3.1&28.3$\pm$1.4&25.7$\pm$2.8&N/A& \textbf{21.4$\pm$1.6} &27.5$\pm$0.4 \\
  &\textbf{Avg. Error} &33.5$\pm$0.3&26.7$\pm$0.7& 23.4$\pm$0.9&25.9$\pm$1.1 &26.7$\pm$0.7 &53.9$\pm$1.7&24.7$\pm$0.7 & 22.3$\pm$1.5 &N/A& \textbf{22.1$\pm$1.2} &23.0$\pm$0.5 \\
  &\textbf{Avg. Runtime} & N/A & 4min& 0.1min& 1min & 4min& 1s&10min&0.5min &N/A& 2min & 0.3min \\
    \hline

\end{tabular}
}
\vspace{-0.0mm}
\label{table:sentiment}
\end{table*}

\begin{table*}[t]
\caption{USPS dataset: classification error of the competing metric learning model. The reported performance is averaged over five random repetitions, and methods with the best performance are marked as bolded black.}
\scriptsize
\centering
\scalebox{1.16}{
\begin{tabular}{c|c||c|c|c|c||c|c||c|c|c}
\hline
{Task} & $\#$Classes&stEuc &stLMNN  & stSCML & stOASIS &mtLMNN & mtSCML & ELLA& Ours+OASIS& Ours+SCML\\
 \hline
 \hline
   \textbf{1}  & 2&0.13 $\pm$0.1&0.09$\pm$0.0& 0.06$\pm$0.0&  0.37$\pm$0.2& 0.09$\pm$0.1&0.09$\pm$0.1 & N/A& 0.22$\pm$0.2 &\textbf{0.03 $\pm$0.1} \\
    \hline
  \textbf{2}  &2 &2.20$\pm$0.8&2.05$\pm$0.6& 2.20$\pm$0.4&  1.86$\pm$0.0& 2.10$\pm$0.6&1.70$\pm$0.1& N/A& \textbf{1.52$\pm$0.6}& 2.12$\pm$0.0 \\
   \hline
  \textbf{3}  & 3& 4.19$\pm$1.0&3.59$\pm$0.8& \textbf{3.18$\pm$0.3}&  5.23$\pm$0.4& 3.92$\pm$1.2& 3.41$\pm$1.3& N/A& 4.58$\pm$0.1& 5.74$\pm$1.4  \\
    \hline
  \textbf{4}  & 3&7.00$\pm$2.0&6.67$\pm$1.6& 6.82$\pm$1.4&  \textbf{ 4.40$\pm$0.2}&6.60$\pm$2.0&5.04$\pm$0.6& N/A&4.43$\pm$1.1 &4.51$\pm$0.3 \\
     \hline
  \textbf{Avg. Error} &N/A &3.37$\pm$0.0 &3.10$\pm$0.0&3.07$\pm$0.1 &2.97$\pm$0.1&3.18$\pm$0.1 & \textbf{2.66$\pm$0.2} & N/A&2.68$\pm$0.1 &3.09 $\pm$0.3 \\
     \hline
\end{tabular}
}
\vspace{-0.0mm}
\label{table:usps}
\end{table*}

\subsection{Real Datasets}
According to whether the label is consistent or not, we categorize the real datasets into two different scenarios: \textbf{label-consistent} and \textbf{label-inconsistent}. In the following, we  will demonstrate the effectiveness of our proposed LML framework in the different datasets.

\begin{table}[t]
\caption{\small Statistics of the \textbf{label-consistent} datasets:}
\scriptsize
\centering
\scalebox{0.98}{
\begin{tabular}{c|c|c|c|c|c}
\hline
{Dataset}&\#Classes&\#Samples & \#Dimension&\#Tasks& Problem Type \\
 \hline\hline
  \centering \textbf{Sentiment} & 2 & 6400& 200 &4& Different Task \\
   \hline
   \centering \textbf{Isolet} & 26 & 7797 & 617 & 5& Same Task \\
   \hline
\end{tabular}
}
\vspace{-1.0pt}
\label{table:consistent_data}
\end{table}

\textbf{Label-consistent datasets:} the label set is shared by all the metric tasks, which can be roughly categorized as: same metric task and different metric tasks with same label set. Therefore, depending on whether is the same task or not, we adopt two datasets in this paper. As shown in Table \ref{table:consistent_data}, \textbf{Sentiment} \cite{blitzer2007biographies} consists of Amazon reviews on four product types (kitchen appliances, DVDs, books and electronics). We randomly split the dataset into training ($800$ samples), validation ($400$ samples) and testing ($400$ samples) sets.  \textbf{Isolet} \footnote{http://www.cad.zju.edu.cn/home/dengcai/Data/MLData.html} dataset, which is a popular dataset for multi-task learning consists of 5 disjoint subjects called isolet 1-5. We randomly split each task of the dataset into training ($10\%$ samples), validation ($20\%$ samples) and testing ($70\%$ samples) sets. Moreover, we set the basis number of \textbf{stSCML} as 100 and 500 in \textbf{Sentiment} and \textbf{Isolet}, respectively.

The experimental results averaged over five random repetitions are presented in Table~\ref{table:sentiment}, and we can conclude that:
\begin{itemize}
  \item Compared with other competing methods, our proposed LML framework outperforms the state-of-the-arts with the average error as $20.3$ and $21.7$ and achieving $2.6\%$ and $0.6\%$ improvement in term of classification error using \textbf{Sentiment} and \textbf{Isolet} datasets, which verifies the effectiveness of our LML framework in a lifelong learning manner. Furthermore, the performance of our LML framework is also better than the existing lifelong learning model (\textbf{ELLA}), due to the fact that we adopt the lifelong dictionary in LML framework, which keeps on learning step-by-steps.
  \item For the real datasets, Table~\ref{table:sentiment} also shows that the comparison of time consumption between our LML framework and other single / multi-task metric models. Our LML framework is more efficient than most state-of-the-arts due to we do not need to retrain all the previous tasks. However, our LML framework is little slower than the multi-task metric model in \textbf{Isolet} and faster than \textbf{LMNN}. This is because we set the high dimensional transformed features.
  \item  Similarity metric function outperforms distance metric function on \textbf{Sentiment} dataset, which implies that similarity metric may be important for different tasks; meanwhile, distance metric outperforms similarity metric on the \textbf{Isolet} dataset, which implies that distance metric may be important for this metric task.
\end{itemize}

\textbf{Label-inconsistent datasets:} the label set of coming metric task is different from the learned metric tasks. \textbf{USPS} \footnote{http://statweb.stanford.edu/~tibs/ElemStatLearn/data.html} dataset consists of 7291 $16\times 16$ gray-scale images of digits $0-9$ automatically. The features are $256$-d grayscale values. We split all the classification problem into 4 tasks:$\{ 0,1\}, \{2,3\}, \{4,5,6\}$ and $\{7,8,9\}$, respectively. Therefore, the number of classes of each task is 2, 2, 3, 3. We use randomly selected $10\%$ of the all the samples as training samples while the remaining for test. From the presented result in Table \ref{table:usps}, we can notice that the performance of our LML framework leads to the second best one with the average error as $2.68$, only $0.02$ worse than the best one \textbf{mtSCML} and outperforms other state-of-the-arts with a big gap. This is because our LML only train the model using the data from the only one corresponding task, instead of \textbf{mtSCML} adopting the data of all tasks together for model training. That is why ours is more efficient than \textbf{mtSCML} as shown in Table~\ref{table:sentiment}.

\subsection{Evaluating Lifelong Metric learning Framework}
In this subsection, we conduct comparisons on the proposed lifelong metric learning formulation, and study how the learned task impact its generalization performance.

\subsubsection{Effect of the $\left\|W\right\|_{1,\mr{off}}$-norm Regularization}
\begin{figure}[t]
\centering
\includegraphics[width =253pt,height =155pt]{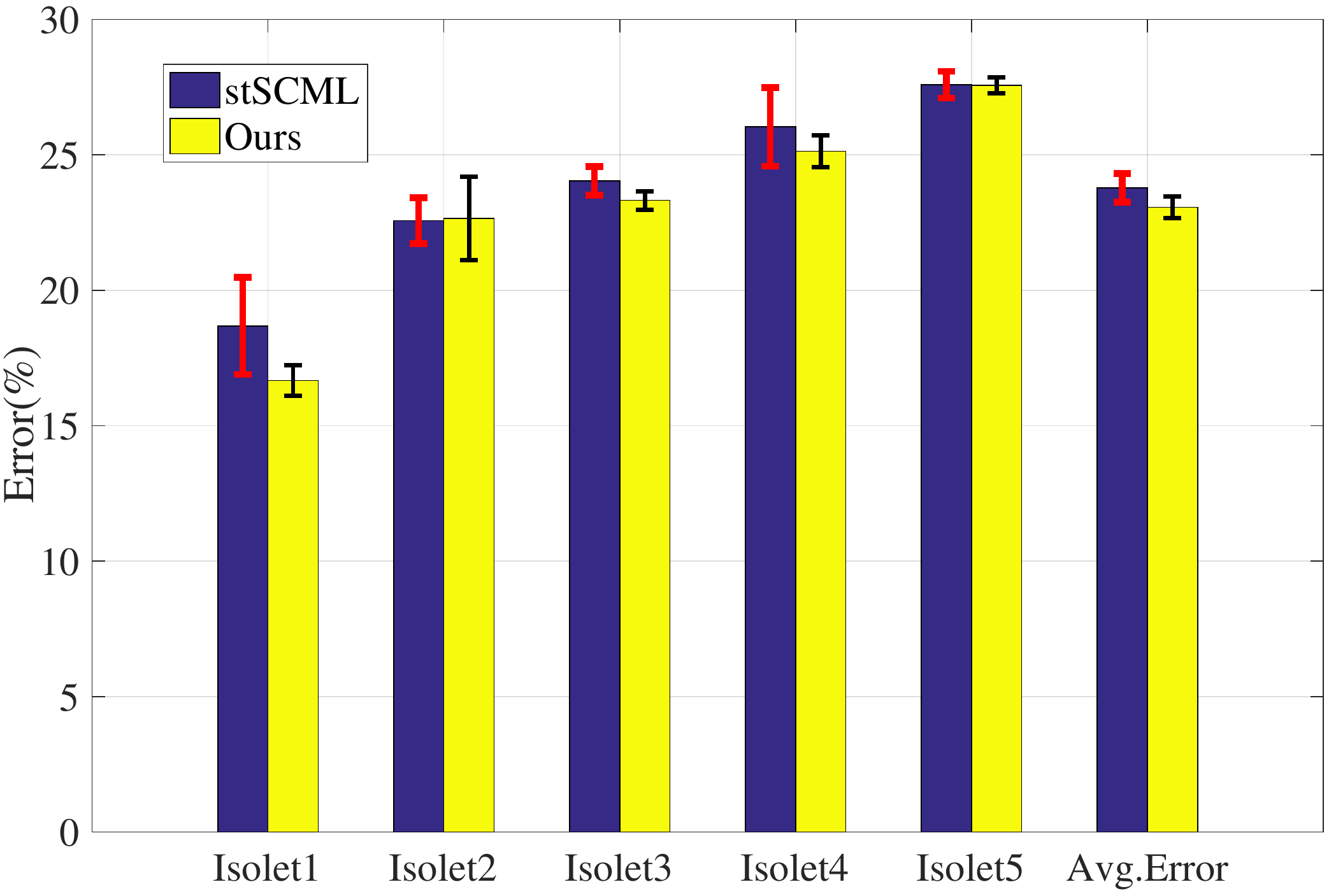}
\vspace{-2.0pt}
\caption{The effect of the $\left\|\cdot\right\|_{1,\mr{off}}$-norm. The horizontal and vertical axes are the index of task and classification error of \textbf{Isolet} dataset, respectively. In addition, the red line and black line are the corresponding standard deviation of \textbf{stSCML} and \textbf{ours}, respectively.}
\label{fig:offnorm}
 \vspace{-1pt}
\end{figure}

In order to study how the $\left\|W\right\|_{1,\mr{off}}$-norm regularization affect the performance of the single metric task, we compare the \textbf{stSCML} method with our proposed framework Eq.~\eqref{eq:model1} on the \textbf{Isolet} dataset. Specifically, we remove the regularization term of $L_0$ in Eq.~\eqref{eq:model1}, i.e., $\gamma\left\|L_0\right\|_F^2$, and employ FISTA~\cite{beck2009fast} to efficiently optimize such a convex problem. We also randomly split each task of the \textbf{Isolet} dataset into training ($10\%$ samples), validation ($20\%$ samples) and testing ($70\%$ samples) sets, and the performance (averaged over 5 random repetitions) is presented in Fig.~\ref{fig:offnorm}. In general, our model in Eq.~\eqref{eq:model1} outperforms \textbf{stSCML} on all the single learned task expect for task \textbf{Isolet2}. This observation verities that the correlation information among different transformed features enables to improve the learning efficacy, i.e., the effectiveness of $\left\|W\right\|_{1,\mr{off}}$-norm.

\subsubsection{Effect of the Dimension of Transformed Features}
\begin{figure}[t]
\centering
\includegraphics[width =253pt,height =155pt]{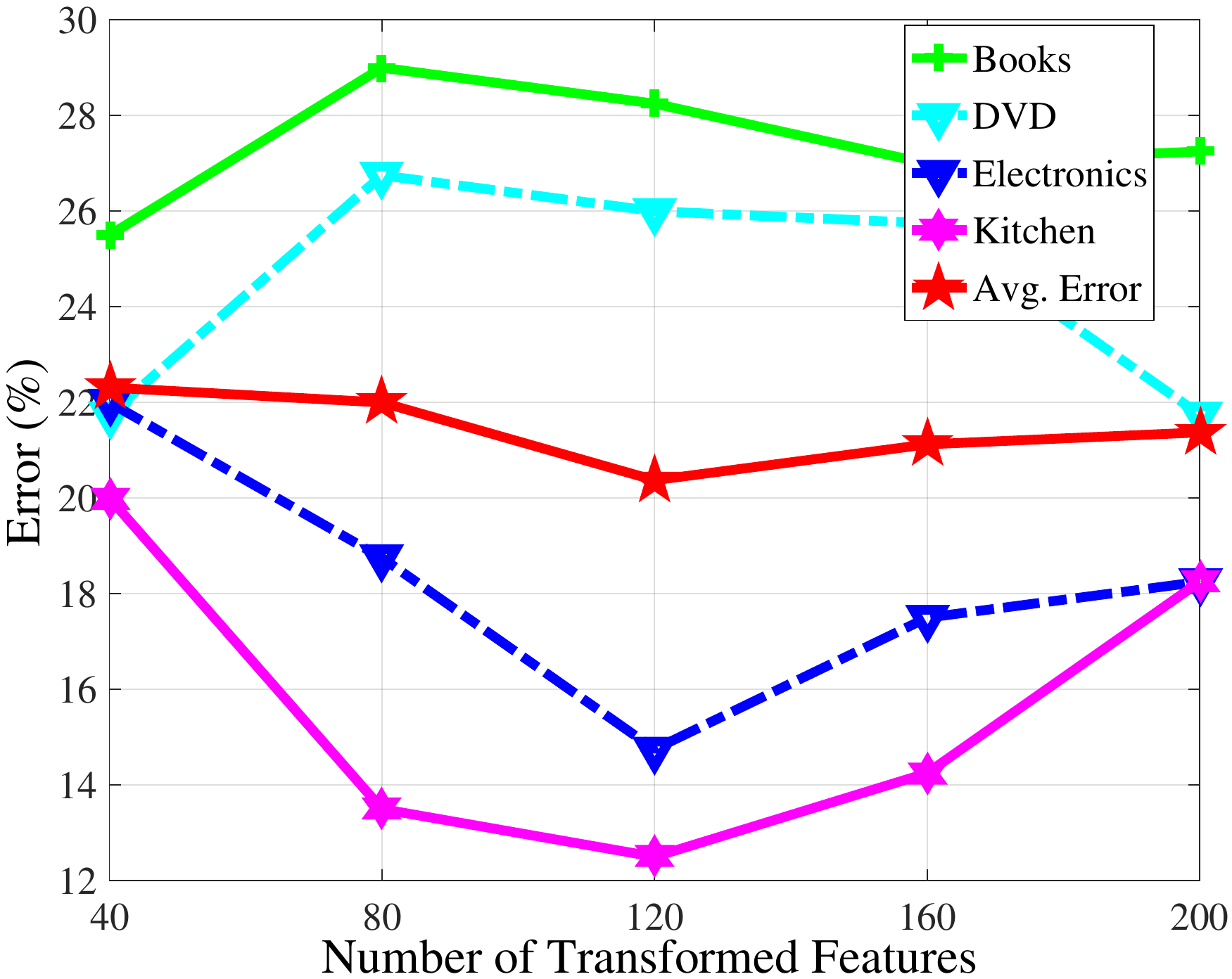}
\vspace{-2.0pt}
\caption{The Effect of the Dimension of Transformed Features $d$. The horizontal and vertical axes are the dimension of transformed features and classification error of \textbf{Sentiment} dataset, respectively.}
\label{fig:numbasis}
 \vspace{-1pt}
\end{figure}

In this subsection, we utilize the \textbf{Sentiment} dataset to evaluate how the dimension of transformed features $d$ affect the performance of our LML framework (Ours+OASIS) in term of classification error. Specifically, we also randomly split the dataset into training (800 samples), validation (400 samples), and testing (400 samples) sets. By varying the number of transformed feature $d$ from 40 to 200, we present the performance (averaged over 5 random repetitions) as shown in Fig.~\ref{fig:numbasis}. Notice that average classification error changes with different number of transformed features, which verifies that all the metric task should be embedded in a low-dimensional subspace, namely lifelong dictionary in our LML framework. In addition, the error of average classification is minimum when $d=120$, i.e., the performance of our LML is best. After that, the classification error is decreasing with the increase of size of $d$. This is because that the larger size of $d$, the more redundant feature information can be involved in the lifelong dictionary $L$.

\subsubsection{Effect of the Number of Learned Tasks}
\begin{figure}[t]
\centering
\includegraphics[width =253pt,height =155pt]{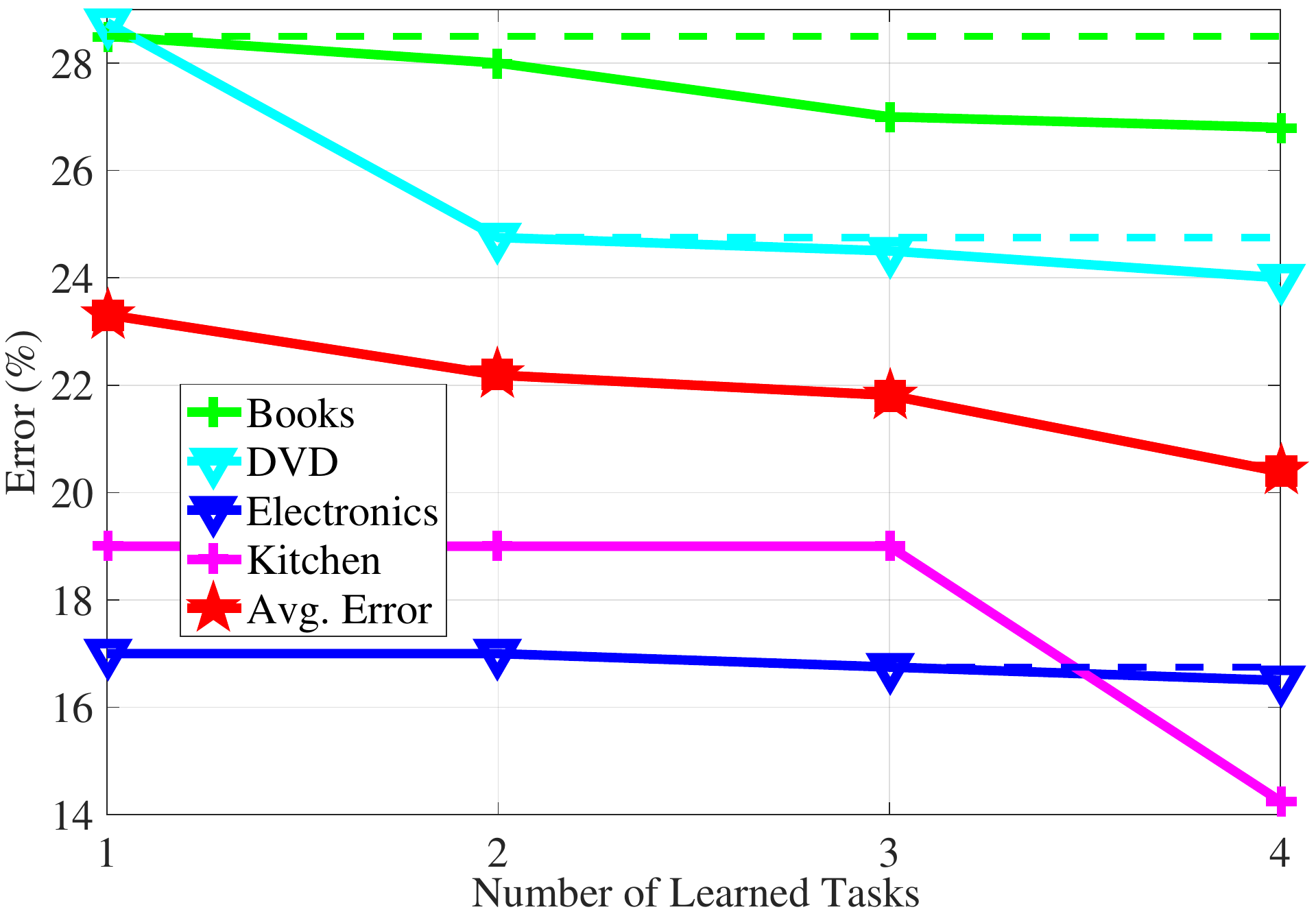}
\caption{The Effect of the Number of Learned Tasks. The horizontal and vertical axes are the number of learned tasks and classification error of each task, respectively. The initial classification error of each task is achieved using sinlge metric learning. Notice that the average error is decreased when increasing the number of tasks}
 \vspace{-2pt}
\label{fig:lifelonglearning}
 \vspace{-1pt}
\end{figure}

In this subsection, we also adopt \textbf{Sentiment} dataset to study how the number of learned tasks $t$ affect the classification performance of our LML framework. After splitting the dataset into training (800 samples), validation (400 samples) and testing (400 samples) sets, we set the sequence of learned $t$ tasks as: Books, DVD, Electronics and Kitchen; we present the classification performance (averaged over 5 random repetitions) in Fig.~\ref{fig:lifelonglearning}. Obviously, as each metric task is imposed step-by-step, the error of our LML is decreased, i.e., the performance of our LML framework is improved gradually, which justifies that our LML framework can accumulate knowledge continuously and achieve lifelong learning like ``human learning''. In addition, the performance of early learned tasks are improved more obviously than succeeding task, i.e., the early tasks can benefit from the accumulated knowledge.

\vspace{-0mm}
\section{CONCLUSION} \label{sec:conclusion}
In this paper, we study how to add metric task into original metric system without retraining the whole system in a too time consuming way as most state-of-the-art online metric learning models. Specifically, we propose lifelong metric learning (LML) framework, which learns ``\textbf{lifelong dictionary}'' as shared basis for all metric models based on the assumption that all metric tasks are retained in a low-dimensional common subspace. When new metric task arrives, our LML can transfer knowledge through the shared lifelong dictionary to learn the new coming metric model with sparsity regularization, and redefine the basis metrics with knowledge from the new metric task. After converting this convex problem into two subproblems via Online Passive Aggressive optimization, we adopt proximal gradient method to solve our proposed LML framework. Through extensive experiments carried our on several multi-task datasets, we verify that our proposed framework are well suited to the lifelong learning problem, and exhibit prominent performance in both effectiveness and efficiency.





\ifCLASSOPTIONcaptionsoff
  \newpage
\fi




\bibliographystyle{IEEEtran}
%

\bibliography{MetricLearning}
\end{document}